%% file: _main.tex
\documentclass[runningheads]{llncs}

\usepackage{eccv}

\usepackage{eccvabbrv}

\usepackage{graphicx}
\usepackage{booktabs}

\usepackage[accsupp]{axessibility}  

\usepackage{hyperref}

\usepackage{orcidlink}

\newlength\savewidth
\newcommand{\tablestyle}[2]{\setlength{\tabcolsep}{#1}\renewcommand{\arraystretch}{#2}\centering\footnotesize}

\begin{document}

\title{LaPose: Laplacian Mixture Shape Modeling for RGB-Based Category-Level Object Pose Estimation} 

\titlerunning{LaPose}

\author{Ruida Zhang\inst{1} \and
Ziqin Huang\inst{1} \and
Gu Wang\inst{1} \and
Chenyangguang Zhang\inst{1} \and
Yan Di\inst{2} \and
Xingxing Zuo\inst{3} \and
Jiwen Tang\inst{1} \and
Xiangyang Ji\inst{1}
}

\authorrunning{R. Zhang \etal}

\institute{Tsinghua University \and
Technical University of Munich \and
California Institute of Technology \\
\email{\{zhangrd23@mails. xyji@\}tsinghua.edu.cn}
}

\maketitle

\input{Sections/0-abstract}

\input{Sections/1-introduction}
\input{Sections/2-related-work}
\input{Sections/3-method}
\input{Sections/4-experiment}
\input{Sections/5-conclusion}
\bibliographystyle{splncs04}
\bibliography{egbib}
\end{document}

%% file: Sections/0-abstract.tex
\begin{abstract}

While RGBD-based methods for category-level object pose estimation hold promise, their reliance on depth data limits their applicability in diverse scenarios. 
In response, recent efforts have turned to RGB-based methods; however, they face significant challenges stemming from the absence of depth information.
On one hand, the lack of depth exacerbates the difficulty in handling intra-class shape variation, resulting in increased uncertainty in shape predictions. 
On the other hand, RGB-only inputs introduce inherent scale ambiguity, rendering the estimation of object size and translation an ill-posed problem.
To tackle these challenges, we propose \textbf{LaPose}, a novel framework that models the object shape as the \textbf{La}placian mixture model for \textbf{Pose} estimation.
By representing each point as a probabilistic distribution, we explicitly quantify the shape uncertainty.
LaPose leverages both a generalized 3D information stream and a specialized feature stream to independently predict the Laplacian distribution for each point, capturing different aspects of object geometry.
These two distributions are then integrated as a Laplacian mixture model to establish the 2D-3D correspondences, which are utilized to solve the pose via the PnP module.
In order to mitigate scale ambiguity, we introduce a scale-agnostic representation for object size and translation, enhancing training efficiency and overall robustness.
Extensive experiments on the NOCS datasets validate the effectiveness of LaPose, yielding state-of-the-art performance in RGB-based category-level object pose estimation.
Codes are released at \url{https://github.com/lolrudy/LaPose}.
  \keywords{Category-level Object Pose Estimation \and 3D Object Detection \and PnP Algorithm}
\end{abstract}

%% file: Sections/1-introduction.tex
\section{Introduction}

\begin{figure}[t]
\centering
\includegraphics[width=0.95\textwidth]{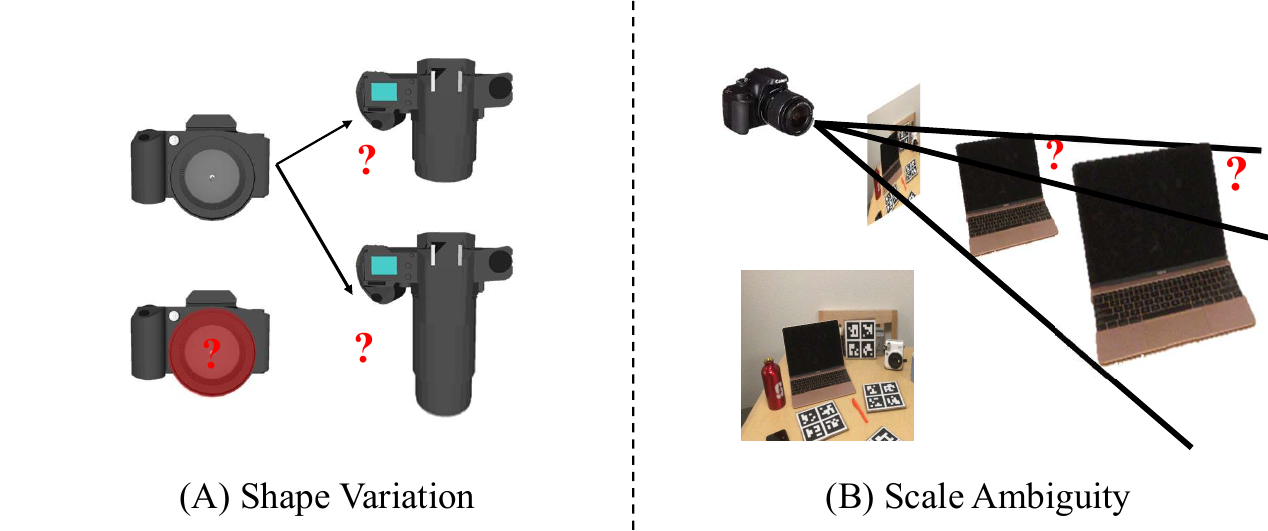}
\caption{Two main challenges of RGB-based category-level object pose estimation.
(A) The lack of depth information exacerbates the difficulty in handling intra-class shape variation.
The length of the camera lens is uncertain in the front view.
(B) The RGB-only inputs introduce scale ambiguity. 
Laptops of various sizes have identical appearance in the image.
}
\label{fig:teaser}
\end{figure}

The task of category-level object pose estimation involves predicting the 9DoF pose, including 3D rotation, 3D translation, and 3D metric size, for unseen objects from a given set of categories.
This field has garnered increasing research interest for its wide applications in robotics~\cite{robotics}, augmented reality (AR), virtual reality (VR)~\cite{arvr}, and 3D understanding~\cite{nie2020total3dunderstanding}.
While RGBD-based methods~\cite{fs-net, GPV-Pose, zhang2022ssp, zhang2022rbp, lin2022dpdn, zheng2023hs, liu2023istnet, lin2023vinet, dualposenet, shape_deform, cass, sgpa, NOCS} demonstrate promising results, most of them heavily rely on depth sensors, which restricts their applicability in general scenarios~\cite{wei2023dmsr, lee2021msos}.

Therefore, RGB-based category-level object pose estimation methods \cite{lee2021msos, chen2020rgbsyn, fan2022oldnet, wei2023dmsr} have been proposed as alternatives suitable for deployment on embedded devices such as AR headsets and mobile phones~\cite{wei2023dmsr}. 
However, as shown in Fig.~\ref{fig:teaser}, the absence of depth information presents two significant challenges: 
firstly, the lack of depth complicates the prediction of object shape and exacerbates the difficulty in handling intra-class shape variation;
secondly, relying solely on the RGB input introduces inherent scale ambiguity, rendering the estimation of translation and size an ill-posed problem.

To address these challenges, two lines of solutions are proposed by recent methods. 
On one hand, MSOS \cite{lee2021msos} and OLD-Net \cite{fan2022oldnet} estimate the metric depth and Normalized Object Coordinate Space (NOCS) coordinates~\cite{NOCS} to establish 3D-3D correspondences and solve the pose via the Umeyama algorithm~\cite{umeyama}.  
On the other hand, DMSR \cite{wei2023dmsr} utilizes object normal and relative depth predicted by pretrained DPT models \cite{Ranftl_2021_ICCV_dpt} as additional inputs to estimate the NOCS coordinate map as well as the object metric scale, and then solves the pose using the PnP algorithm~\cite{lepetit2009epnp}. 

However, these methods encounter limitations in two key aspects.
Firstly, as depicted in \cref{fig:teaser} (A), the absence of depth information raises challenges in accurately measuring the object shape.
The shape uncertainty is particularly evident in certain areas of the image (such as the camera lens in \cref{fig:teaser} (A)), complicating the establishment of precise correspondences.
All these methods~\cite{lee2021msos,fan2022oldnet,wei2023dmsr} treat the predicted correspondences of each pixel equally and rely on RANSAC to filter outliers, which slows down the prediction process and undermines robustness. 
Secondly, MSOS \cite{lee2021msos} and OLD-Net \cite{fan2022oldnet} do not consider scale ambiguity. 
In contrast, DMSR \cite{wei2023dmsr} utilizes the same features to predict the NOCS map and the metric scale.
However, inferring the metric scale from a single RGB image is inherently ill-posed, potentially leading to unstable training of other components and inferior results.

To tackle these problems,
we propose \textbf{LaPose} by modelling object shape as \textbf{La}placian mixture model for RGB-based category-level object \textbf{Pose} estimation.
As shown in Fig.~\ref{fig:teaser}, the shape uncertainty of each pixel varies due to the lack of depth. 
In comparison to predicting NOCS map deterministically as the previous methods, modeling the NOCS coordinate as the probabilistic distribution introduces additional information of its variance, which explicitly measures the shape uncertainty of each point.

In order to provide a more comprehensive understanding of the object shape, we employ the Laplacian Mixture Model (LMM), which combines two independent Laplacian distributions inferred from different sources of information.
We choose the Laplacian distribution due to its superior ability to handle outliers compared to other distributions such as Gaussian \cite{chen2020monopair}.
We utilize two parallel information streams to capture different aspects of the object geometry and independently predict the Laplacian distribution of the NOCS map.
To extract a generalizable category-agnostic 3D feature, we leverage DINOv2 as the generalized 3D information stream.
Several studies~\cite{zhang2023tale, amir2022dino, chen2023secondpose} have demonstrated DINOv2's capability of encapsulating 3D information.
Specifically, it can extract SE(3)-consistent local features to establish semantic correspondences across objects of varying shapes and poses \cite{chen2023secondpose}.
However, as DINOv2 is trained in a category-agnostic manner, it is inferior to extract category-specific features.
To complement this, we train another backbone~\cite{liu2022convnext} dedicated to capturing category-specific information as the specialized feature stream.

After estimating the Laplacian mixture model, we establish the 2D-3D correspondences and solve the pose utilizing the PnP module constructed by a convolutional network similar to \cite{GDRN}.
The PnP module benefits from LMM in two key aspects: 
firstly, it can dynamically aggregate diverse object geometry information captured by the dual information branch; 
secondly, it can identify the shape uncertainty in different areas and filter out erroneous correspondences, thereby enhancing overall robustness. 
This novel integration of LMM facilitates more informed decision-making during pose estimation, contributing to the model's reliability in challenging scenarios.

Moreover, to address the challenge of inherent scale ambiguity in pose prediction, we propose a scale-agnostic representation for both translation and size. 
By decoupling the pose prediction from the metric scale, our method enhances training efficiency by cutting off the propagation of errors resulting from scale ambiguity to pose estimation. 
Moreover, this approach leads to more informative evaluation metrics, providing deeper insights into the performance of the pose estimation methods across various object scales.

In summary, our contributions are three-fold:
\begin{itemize}
    \item We introduce the Laplacian Mixture Model (LMM) to effectively model the object shape, explicitly quantifying shape uncertainty for each point and addressing intra-class shape variation.
    \item  We propose a dual-stream framework to estimate the parameters of LMM, which essentially harnesses a generalized 3D information stream and a specialized feature stream to capture diverse aspects of object geometry.
    \item We propose a scale-agnostic 9DoF pose representation, improving training efficiency and providing informative evaluation metrics.
\end{itemize}

LaPose achieves state-of-the-art performance on NOCS datasets \cite{NOCS}, with extensive experiments conducted to demonstrate the effectiveness of our design choices.

%% file: Sections/2-related-work.tex
\section{Related Work}

\subsection{Instance-level Object Pose Estimation}

Instance-level object pose estimation methods aim to predict the 6DoF object pose using either RGB or RGBD data. 
For RGB-based methods, some directly regress object poses from input image \cite{kehl2017ssd,xiang2017posecnn,labbe2020cosypose,li2019deepim}, while others establish 2D-3D correspondences through keypoint detection or pixel-wise 3D coordinate estimation and then employ PnP algorithms to solve the pose~\cite{hodan2020epos, peng2019pvnet, park2019pix2pose, hybridpose, zakharov2019dpod, su2022zebrapose, chen2022epro}. 
While direct methods tend to be faster, correspondence-based methods offer higher accuracy at the expense of increased computational complexity. 
Recent works \cite{sopose,li2019cdpn,GDRN} aim to address these trade-offs by combining the advantages of both approaches, pursuing end-to-end architecture and real-time performance.
As for RGBD-based methods, \cite{FFB6D,pvn3d,wang2019densefusion,kehl2016deep,Wohlhart2015b, lipson2022coupled, hu2022perspective} leverage both the RGB image and observed depth to predict 6D poses.
Despite achieving promising performance on common datasets, the practical applications of instance-level pose estimation are hindered, as they can only handle a small number of objects and require accurate CAD models of the target objects.

\subsection{Category-level object pose estimation}

Category-level object pose estimation is essential for real-world applications as it eliminates the need for precise CAD models for individual object instances. 
The concept of Normalized Object Coordinate Space (NOCS) \cite{NOCS} offers a unified coordinate representation for object instances within a category.
While some subsequent methods \cite{shape_deform,sgpa,donet,ACR-Pose,lin2022dpdn} estimate the pose by predicting NOCS coordinates for each point and establishing 3D-3D correspondences,
\cite{dualposenet,GPV-Pose,lin2023vinet,fs-net,zhang2022ssp,zhang2022rbp} directly regress the 9DoF pose for efficiency.
While these methods demonstrate promising results, their applicability is constrained by their heavy reliance on the observed depth.

As for RGB-only methods, challenges arise due to the inherent difficulty in recovering 3D information from a single RGB image and the presence of scale ambiguity. 
Recent efforts have aimed to overcome these limitations. 
Exemplarily, Chen \etal \cite{chen2020rgbsyn} introduces an analysis-by-synthesis framework by combining gradient-based fitting with neural image synthesis.
MSOS \cite{lee2021msos} and OLD-Net \cite{fan2022oldnet} predict the pose via Umeyama algorithm \cite{umeyama} from the 3D-3D correspondences established by NOCS coordinates prediction and metric depth estimation.
DMSR \cite{wei2023dmsr} uses object normals and relative depths estimated by pretrained DPT model \cite{Ranftl_2021_ICCV_dpt} as additional inputs and estimates the pose from the 2D-3D correspondences via PnP algorithm \cite{lepetit2009epnp}. 
While these methods represent significant advancements in RGB-based category-level pose estimation, the robustness and accuracy remain to be enhanced.

%% file: Sections/3-method.tex
\section{Method}
In this paper, our goal is to solve the problem of RGB-based category-level object pose estimation. Specifically, given an RGB image containing objects from a predefined set of categories, our objective is to detect all instances of objects present in the scene and accurately estimate their 9DoF poses. The 9DoF pose includes the 3DoF rotation $\mathbf{R} \in SO(3)$, the 3DoF translation $\mathbf{t} \in \mathbb{R}^{3}$ and 3DoF size $\mathbf{s} \in \mathbb{R}^{3}$.

\input{figs/pipeline-ddpnp}

To this end, we present \textbf{LaPose}, a novel approach that models object shape using the \textbf{La}placian mixture model for \textbf{Pose} estimation (\cref{fig:method}).
We first adopt an off-the-shelf object detector MaskRCNN~\cite{maskrcnn} to crop the object of interest as input.
Then we estimate the parameters of LMM by employing two information streams to predict two Laplacian distributions $Laplace(\mu_{dino}, \sigma_{dino}^2)$, $Laplace(\mu_{conv}, \sigma_{conv}^2)$ of the NOCS coordinate map independently(\cref{sec:dual-stream}).
The estimated  LMM is then utilized for pose estimation through LMM-based PnP solving (\cref{sec:uncertainty}).
Finally, the scale-agnostic pose $\{\mathbf{R}, \mathbf{t}_{norm}, \mathbf{s}_{norm}\}$ is calculated by the PnP module (\cref{sec:scale}).
The overview of LaPose is shown in \cref{fig:method} and we detail each component in the following sections.

\subsection{Dual-Stream LMM Modeling \label{sec:dual-stream}}
\subsubsection{Generalized 3D information stream.} 
DINOv2 \cite{oquab2023dinov2}, as an outstanding foundation model, has demonstrated the ability to establish zero-shot semantic correspondences across images and extract rich 3D information from images~\cite{zhang2023tale, amir2022dino, chen2023secondpose}. 
Specifically, DINOv2 can provide SE(3)-consistent patch-wise local features \cite{chen2023secondpose}, which aligns with the SE(3)-invariant nature of NOCS coordinates, thereby facilitating the learning process of the NOCS map.
Hence, we integrate pretrained DINOv2 into our framework as the generalized 3D information stream to predict the Laplacian distribution of the NOCS map, enabling the extraction of category-agnostic 3D feature $\mathcal{F}_{dino}$.

\subsubsection{Specialized feature stream.} 
However, solely depending on DINOv2 is insufficient, as we observed in the experiments (\cref{tab:ablation} (C)).
We argue that the category-agnostic training approach leaves DINOv2 lacking in category-specific knowledge essential for pose estimation. 
This observation underscores the need for complementary approaches to enrich the information regarding object geometry.
Therefore, in addition to leveraging the generalized 3D information stream, we further train a convolutional network \cite{liu2022convnext} to extract category-specific features $\mathcal{F}_{conv}$.

\subsubsection{Estimating Parameters of LMM.}
In order to estimate a single Laplacian distribution $Laplace(\mu, \sigma^2)$, we utilize $\mathcal{F}_{dino}$ and $\mathcal{F}_{conv}$ to estimate the pixel-wise mean $\mu$ and variance $\sigma^2$ independently.
We employ the Laplacian aleatoric uncertainty loss from~\cite{gaussianuncertainties, chen2020monopair} to learn both $\mu$ and $\sigma^2$ simultaneously,

\begin{equation}
\begin{split}
   \mathcal{L}_{3D-dino} = 
    \frac{\lambda_1}{\sigma_{dino}^2} ||\mathbf{M}_{vis} \cdot (\mathbf{C}^{3D}_{gt} - \mu_{dino})||_1 + \mathbf{M}_{vis} \cdot \log(\sigma_{dino}^2), \\
    \mathcal{L}_{3D-conv} = 
    \frac{\lambda_2}{\sigma_{conv}^2} ||\mathbf{M}_{vis} \cdot (\mathbf{C}^{3D}_{gt} - \mu_{conv})||_1 + \mathbf{M}_{vis} \cdot \log(\sigma_{conv}^2),
\end{split}
\end{equation}
where $||\boldsymbol{\cdot}||_1$ denotes the L1 distance function, $\mathbf{M}_{vis}$ is the visible mask, $\cdot$ denotes element-wise multiplication, $\mathbf{C}^{3D}_{gt}$ is the ground-truth NOCS map and $\lambda_1, \lambda_2$ are pre-defined hyper-parameters.

When the NOCS coordinate error of $\mu$ is large, the L1-distance term dominates and $\sigma^2$ is forced to be large to reduce this term; 
when the coordinate error is small, $\sigma^2$ is encouraged to be small to reduce the logarithm term.
Therefore, $\sigma_{dino}^2, \sigma_{conv}^2$ are learned in a self-supervised manner.


Finally, the LMM is obtained by combining both estimated distributions $Laplace(\mu_{dino}, \sigma_{dino}^2)$ and $Laplace(\mu_{conv}, $
$\sigma_{conv}^2)$.

\subsection{LMM-based PnP Solving \label{sec:uncertainty}}
We use the estimated LMM to establish 2D-3D correspondences for PnP solving.
We define the optimization goal with L1 re-projection error as follows,
\begin{equation}
    \mathbf{P}_{out} = \mathop{\arg\min}_{\mathbf{P} \in SE(3)} 
 \frac{1}{N} \sum_{i=1}^N {\mathbb{E}}_{\mathbf{x}_i \sim \eta_i} ( || \pi(\mathbf{K} \mathbf{P} \mathbf{x}_i) - \mathbf{u}_i ||_1 ),
    \label{eq:pnp}
\end{equation}
where $\mathbf{u}_i$ is a 2D object point and $N$ is the total number of object points. 
$\eta_i$ is the estimated NOCS coordinate distribution corresponding to $\mathbf{u}_i$ and $\mathbf{x}_i$ is the 3D NOCS point sampled from the distribution. 
$\pi$ is the projection function and $\mathbf{K}$ is the camera intrinsic matrix.
$\mathbf{P}$ is an arbitrary pose in the SE(3) space.

Since directly solving \cref{eq:pnp} is intractable, we adopt a convolutional network $\Phi$ as the PnP module similar to \cite{GDRN} for approximating \cref{eq:pnp}, which enables end-to-end training and efficient inference.
$\Phi$ takes $\mu_{dino}$, $\sigma_{dino}^2$, $\mu_{conv}$, $\sigma_{conv}^2$ and the 2D pixel coordinate map $\mathbf{C}^{2D}$ as input and predicts the scale-agnostic parameters of rotation $\mathbf{R}_{out}$ and translation $\mathbf{t}_{out}$ as
\begin{equation}
    \mathbf{R}_{out}, \mathbf{t}_{out} = \Phi(\mu_{dino}, \sigma_{dino}^2, \mu_{conv},  \sigma_{conv}^2, \mathbf{C}^{2D}).
\end{equation}
The scale-agnostic pose representation $\mathbf{R}_{out}, \mathbf{t}_{out}$ will be detailed in \cref{sec:scale}.


It is notable that the PnP module derives two key benefits from LMM modeling:
Firstly, it dynamically aggregates diverse object geometry information captured by the dual information branches.
Secondly, it effectively discerns shape uncertainty across different regions, facilitating the filtering of erroneous correspondences and enhancing overall robustness.
The integration of LMM facilitates more informed decision-making during pose estimation, contributing to the model's reliability in challenging scenarios.

Meanwhile, since the NOCS maps might not contain full information of object size, we employ a separate size head to predict the scale-agnostic object size parameter $\mathbf{s}_{out}$ (\cref{sec:scale}) from the concatenation of global features of both backbones.

\begin{figure}[t]
\centering
\includegraphics[width=0.8\textwidth]{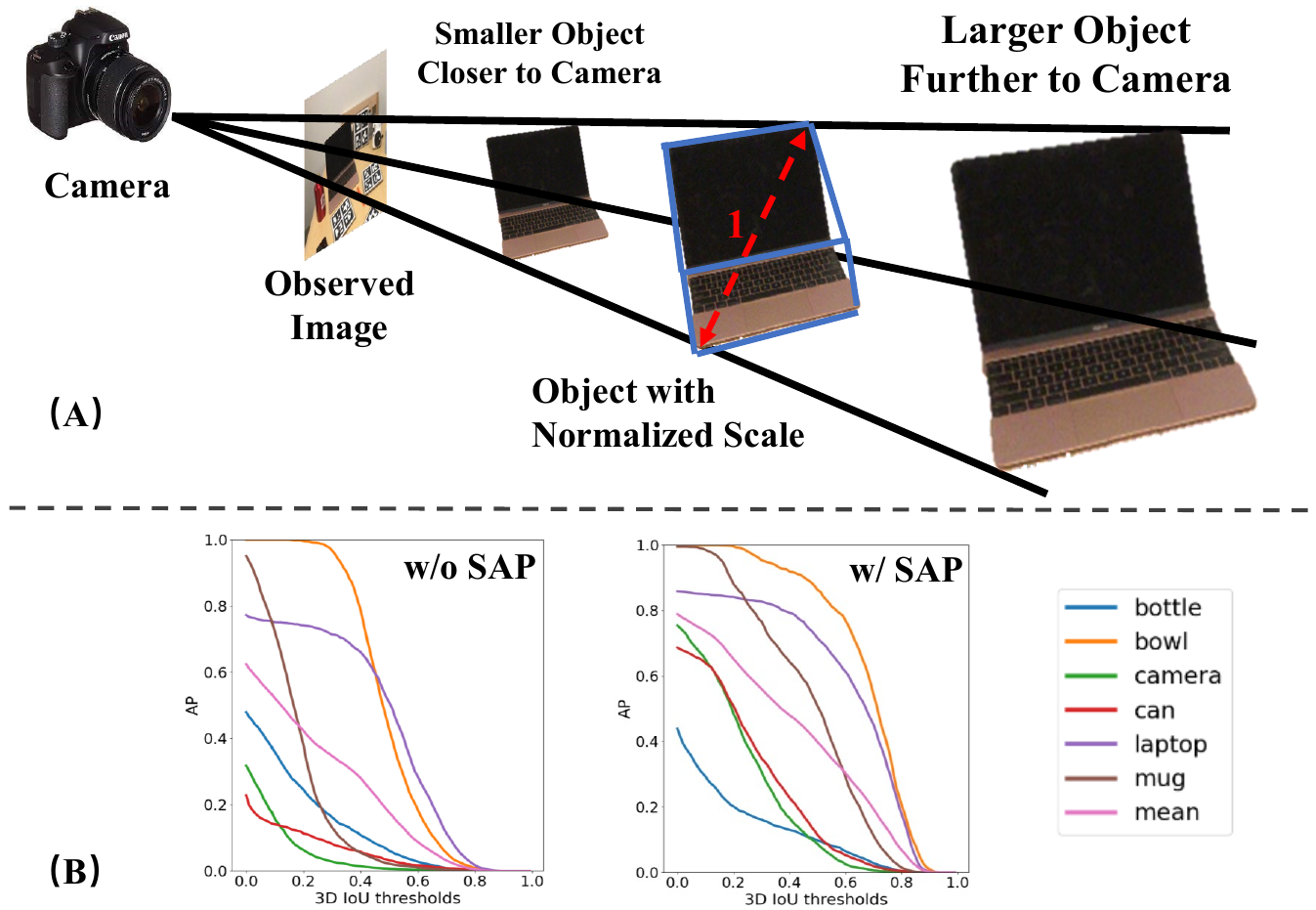}
\caption{\textbf{(A)} Illustration of scale ambiguity: Objects of various scales exhibit identical appearances in the image.
We propose Scale-Agnostic Pose representation (SAP) by normalizing the scale such that the diagonal length of the object tight bounding box is 1. \textbf{(B)} Average Precision on 3D IoU under different thresholds with or without SAP.
}
\label{fig:scale-ambiguity}
\end{figure}

\subsection{Scale-Agnostic Pose Representations \label{sec:scale}}
As shown in \cref{fig:scale-ambiguity} (A), 
the object metric scale cannot be determined solely from a single RGB image.
For instance, consider two scenarios: a larger object positioned further away from the camera and a smaller object located closer to the camera. 
Despite their contrasting sizes, these objects may appear identical in the image, leading to scale ambiguity. 
The lack of depth information makes predicting object translation and size solely depending on visual appearance an ill-posed problem. 
Therefore, we propose a scale-agnostic pose representation,  wherein we normalize the object to fit within a tight bounding box with a diagonal length equal to 1.
Fig.~\ref{fig:scale-ambiguity} (B) shows that adopting the proposed Scale-Agnostic Pose representation (SAP) boosts the performance significantly.

Specifically, for an object with size $\mathbf{s}=\{s_x, s_y, s_z\}$, the normalized size is 
\begin{equation}
    \mathbf{s}_{norm} = \left\{\frac{s_x}{d}, \frac{s_y}{d}, \frac{s_z}{d}\right\},
\end{equation}
where $d=\sqrt{s_x^2+s_y^2+s_z^2}$ is the original diagonal length of the object's tight bounding box.
We compute the average normalized size of the category $\mathbf{s}_{avg}$ beforehand and predict the delta value $\mathbf{s}_{out} = \mathbf{s}_{norm} - \mathbf{s}_{avg}$.

Corresponding to the object size, the translation $\mathbf{t}=\{t_x, t_y, t_z\}$ is also normalized with $d$,
\begin{equation}
    \mathbf{t}_{norm} = \left\{\frac{t_x}{d}, \frac{t_y}{d}, \frac{t_z}{d}\right\} = \{t_x^{(norm)}, t_y^{(norm)}, t_z^{(norm)}\}.
\end{equation}

As in previous RGB-based works \cite{li2019cdpn, GDRN, li2019deepim}, we separate the 3D translation into two components: the 2D location $(o_x, o_y)$ of the
projected object center and the object’s normalized distance $t_z^{(norm)}$ from the camera.


Since the input image is cropped and resized based on the detection results, we regress the detection-invariant translation parameters $\mathbf{t}_{\text{out}} = [\delta_x, \delta_y, \delta_z]$,
\begin{equation}
\begin{cases}
\delta_x =  (o_x - c_x) / w_{box} \\
\delta_y =  (o_y - c_y) / h_{box} \\
\delta_z =  t_z^{(norm)} s_{box} / s_{in}
\end{cases},
\label{eq:t_site}
\end{equation}
where $w_{box}$ and $h_{box}$ represent the width and height of the detected 2D bounding box respectively, $s_{box}=\max(w_{box}, h_{box})$ denotes its size, and $(c_x, c_y)$ indicates its center. 
Additionally, $s_{in}=\max(w_{in}, h_{in})$ denotes the size of the resized input image.


During inference,  $(o_x, o_y, t_z^{(norm)})$ is recovered by
\begin{equation}
\begin{cases}
o_x = \delta_x w_{box} + c_x \\
o_y = \delta_y h_{box} + c_y \\
t_z^{(norm)} = \delta_z s_{in} / s_{box}
\end{cases}.
\label{eq:t_recover}
\end{equation}
The normalized 3D translation is calculated via back-projection given the camera intrinsic matrix $\mathbf{K}$,
\begin{equation}
    \mathbf{t}_{norm} = \mathbf{K}^{-1} t_z^{(norm)} [o_x, o_y, 1].
    \label{eq:t_bp}
\end{equation}

Regarding rotation $\mathbf{R}$, we adopt the same representation method as described in \cite{zhang2022ssp}.
We predict the two rotation vectors along the $x$ and $y$ axes $\mathbf{R}_{out}=[r_x, r_y]$ ($r_x, r_y \in \mathbb{R}^3$), which correspond to the first two columns of the rotation matrix. 
In cases involving objects with rotational symmetry such as \textit{bowl} and \textit{can}, which exhibit symmetry around a particular axis, the rotation vector $r_x$ introduces ambiguity. 
Consequently, only the rotation vector $r_y$ is supervised during training in such scenarios.

In order to predict the object pose with absolute scale, we train a MobileNet \cite{mobilenetv3} to predict the diagonal length $d$ of the object's tight 3D bounding box independently.

\subsection{Overall Training Objective}
The overall training objective $\mathcal{L}$ is defined as
\begin{equation}
    \mathcal{L}=\lambda_{pose} \mathcal{L}_{pose}+\lambda_{3D}(\mathcal{L}_{3D-dino}+\mathcal{L}_{3D-coor}),
\end{equation}
where $\mathcal{L}_{pose}$ includes all loss functions to supervise the learning of scale-invariant 9DoF pose parameters (see Sup. Mat. for details), $\mathcal{L}_{3D-dino}, \mathcal{L}_{3D-coor}$ are introduced in Sec.~\ref{sec:uncertainty}.

%% file: figs/pipeline-ddpnp.tex
\begin{figure}[t]
\centering
\includegraphics[width=0.99\textwidth]{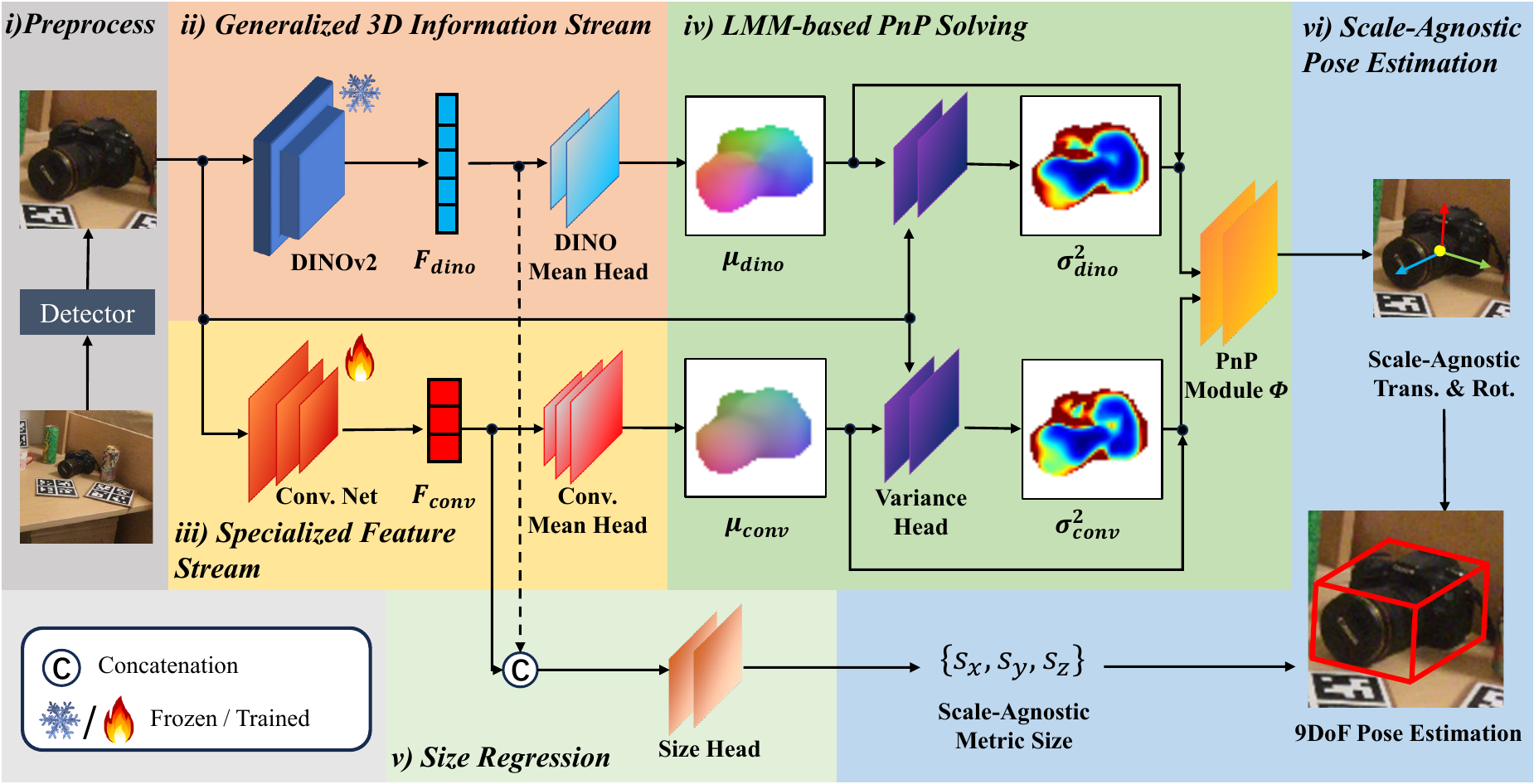}
\caption{Method overview.
i) Given an RGB image, we adopt a detector to crop the object of interest.
The input image is then processed by ii) generalized 3D information stream supported by DINOv2 and iii) specialized feature stream utilizing a convolutional network to extract features $\mathcal{F}_{dino}, \mathcal{F}_{conv}$.
iv) The Laplacian mixture model of the NOCS coordinate map is obtained by combining the Laplacian distributions $Laplace(\mu_{dino}, \sigma_{dino}^2)$ and $ Laplace(\mu_{conv}, \sigma_{conv}^2)$ predicted by both streams.
The subsequent PnP module $\Phi$ solves the translation and rotation from 2D-3D correspondences established by the Laplacian mixture model.
v) Meanwhile, the size head takes $\mathcal{F}_{dino}, \mathcal{F}_{conv}$ as input and predicts the object size.
vi) Finally, the scale-agnostic 9DoF pose parameters are obtained.
}
\label{fig:method}
\end{figure}

%% file: Sections/4-experiment.tex
\section{Experiments}

\subsubsection{Datasets.}
We evaluate the performance of LaPose on the synthetic NOCS-CAMERA25 dataset~\cite{NOCS} and the real-world NOCS-REAL275 dataset~\cite{NOCS} for RGB-based category-level pose estimation.
NOCS-CAMERA25 comprises 300K synthetic images featuring objects rendered onto virtual backgrounds. Of these, 25K images are reserved for testing. 
The dataset encompasses objects from six categories: \textit{bottle, bowl, camera, can, laptop}, and \textit{mug}.
NOCS-REAL275 presents a more challenging real-world dataset, featuring 13 diverse scenes. 
Seven scenes, totaling 4.3K images, are designated for training, while the remaining six scenes, containing 2.7K images, are reserved for testing. 
REAL275 includes objects from the same categories as in CAMERA25.

\subsubsection{Metrics.}
In order to remove scale ambiguity in the evaluation metrics, we use the proposed scale-agnostic pose representation to compute scale-agnostic evaluation metrics.
Therefore, we use the scale-normalized pose to compute the mean average precision of the Normalized 3D Intersection over Union (NIoU) metric at thresholds 25\%, 50\%, and 75\%.
Additionally, we introduce $10^\circ 0.2d$ and $10^\circ 0.5d$ metrics, where an object pose is deemed correct if both its rotation error and normalized translation error fall below the specified threshold.
Specifically, the normalized translation error is presented as the ratio of the diagonal length $d$ of the tight object bounding box.
Moreover, we present $0.2d$, $0.5d$, and $10^\circ$ metrics to assess translation and rotation individually. 
Furthermore, we report results under previously utilized metrics which considers the absolute object scale.
Following \cite{fan2022oldnet, lee2021msos, wei2023dmsr}, we provide the mean Average Precision of $IoU_{25}$, $IoU_{50}$, $IoU_{75}$, $10^\circ 10cm$ and $10cm$.
Since we identified errors in the previous evaluation script given by \cite{NOCS}  (see Sup. Mat.), we have re-evaluated all competitors using the corrected code.

\subsubsection{Implementation Details.}
For the evaluation on REAL275, we train LaPose using the combination of CAMERA25 and REAL275 as \cite{NOCS}.
For the evaluation on CAMERA25, we only use the training data of CAMERA25 as in \cite{fan2022oldnet,wei2023dmsr}.
We employ the MaskRCNN \cite{maskrcnn} as the detector and use detection results generated by \cite{dualposenet} for fair comparison.
All experiments are performed on a single NVIDIA RTX 3090 GPU.
We adopt Ranger optimizer \cite{ranger1,ranger2,ranger3} with the initial learning rate of $10^{-3}$ and the batch size of 32.
The learning rate is annealed at 72\% of the training phase using
a cosine schedule.
We set hyper-parameters $\{\lambda_1, \lambda_2, \lambda_{pose},\lambda_{3D}\}$ as $\{15, 15, 1, 0.1\}$.
We adopt Dynamic Zoom-In \cite{li2019cdpn} during training to make the pose estimation robust to detection errors.
We train a single model for all 6 categories for 100 epochs for \textbf{Ours}, while \textbf{Ours (M)} is a multi-model version where a separate model is trained for each category for 150 epochs.
The inference speed of LaPose is around 10 FPS.

\subsection{Comparison With State-of-the-Art Methods}

\input{tables/norm_result}

\input{tables/abs_result}

\begin{figure}[t]
\centering
\includegraphics[width=0.95\textwidth]{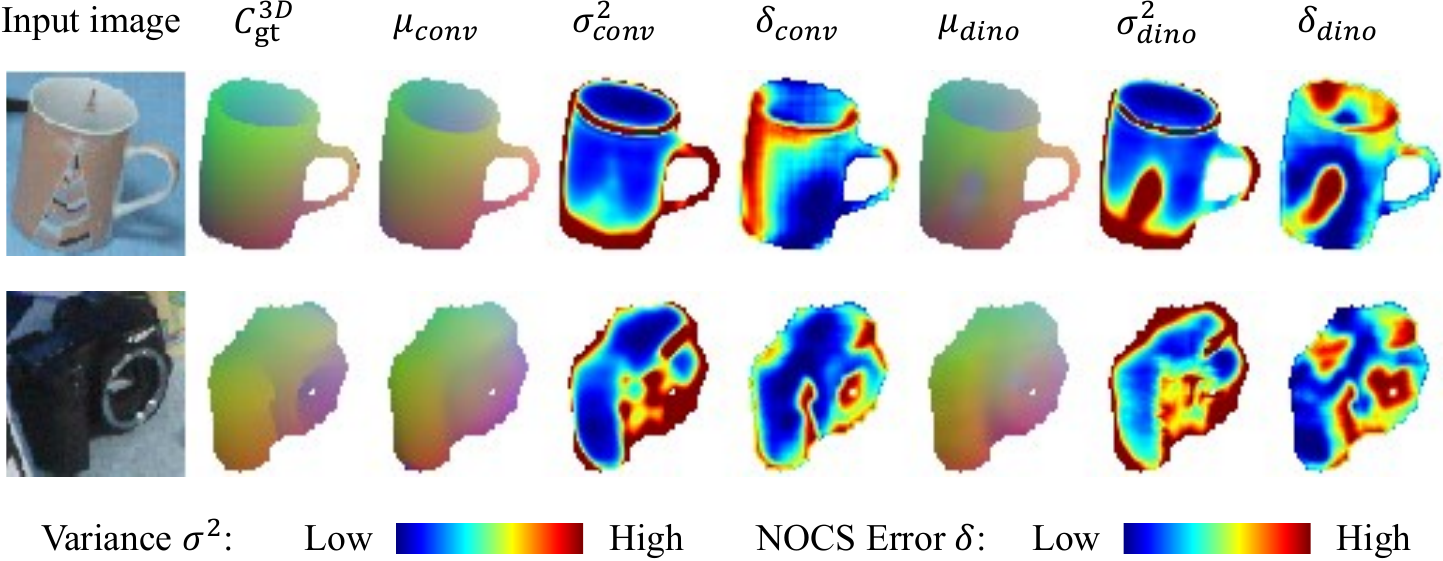}
\caption{Visualization of the predicted Laplacian distribution means $\mu$ and variances $\sigma^2$.
In regions where NOCS errors are pronounced, the variance $\sigma^2$ tends to be higher.
}
\label{fig:vis_uncertainty}
\end{figure}

\begin{figure}[t]
\centering
\includegraphics[width=0.95\textwidth]{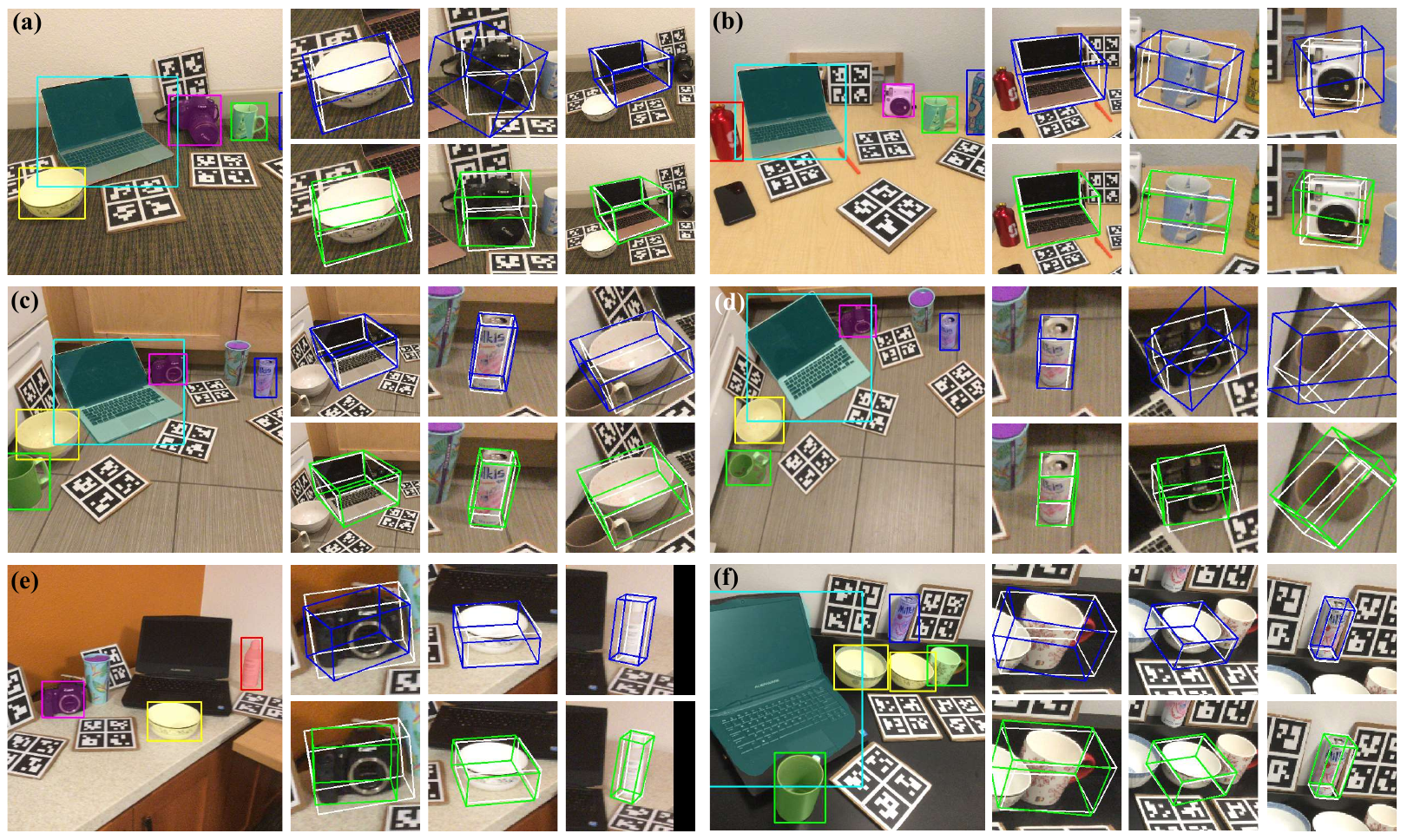}
\caption{Qualitative results of LaPose (green line) and DMSR (blue line) on NOCS-REAL275. Images (a)-(f) demonstrate 2D segmentation results.
}
\label{fig:vis_bbox}
\end{figure}

\input{figs/cat-result-ddpnp}

\input{tables/camera_result}

Tab.~\ref{tab:norm} and Tab.~\ref{tab:abs} present a comprehensive comparison between LaPose and state-of-the-art methods, including MSOS \cite{lee2021msos}, OLD-Net \cite{fan2022oldnet} and DMSR \cite{wei2023dmsr} on NOCS-REAL275 under scale-agnostic and absolute-scale evaluation metrics respectively.

In terms of scale-agnostic evaluation metrics (Tab.~\ref{tab:norm}), LaPose consistently outperforms all competitors by a significant margin across all metrics, demonstrating the effectiveness of our design. 
Specifically, LaPose ({Ours}) surpasses the second-best method DMSR in $NIoU_{25}$ and $10^\circ 0.5d$ by $13.5\%$ and $12.5\%$, respectively. 
Ours (M) exceeds DMSR under the strictest metric $NIoU_{75}$ and $10^\circ 0.2d$ by $10.8\%$ and $13.7\%$, respectively.
In comparison to OLD-Net, Ours achieves an accuracy of 37.4\% in $10^\circ 0.2d$, which is 34.6\% greater than OLD-Net's 2.8\%.

As for evaluation metrics considering the absolute object scale (Tab.~\ref{tab:abs}), Ours (M) also outperforms other competitors under all metrics. 
Specifically, it surpasses the second-best DMSR by $3.2\%$ under $IoU_{25}$, $2.0\%$ under $IoU_{50}$ and $0.9\%$ in $IoU_{75}$. 
Moreover, {Ours} exceeds DMSR and OLD-Net under $10^\circ 10cm$ by $5.3\%$ and $21.9\%$, respectively.

As shown in Fig.~\ref{fig:vis_bbox}, LaPose performs better than DMSR in handling category with large shape variation such as \textit{camera}, thanks to our LMM shape modeling.
Fig.~\ref{fig:category_curve} also proves that our method surpasses DMSR by a large margin in handling category \textit{camera} (green lines in Fig.~\ref{fig:category_curve}).
Additionally, it is noteworthy that LaPose exhibits superior ability in predicting rotations, achieving significantly higher accuracy under small thresholds (as observed in the middle of Fig.\ref{fig:category_curve}).

Fig.~\ref{fig:vis_uncertainty} proves the efficacy of our LMM modeling approach.
In the area where the NOCS prediction error is high, the variance is also correspondingly higher compared to other areas.
This observation demonstrates how our LMM modeling effectively guides the PnP module to prioritize regions with lower variance and more accurate NOCS predictions. 
Consequently, this capability enables LaPose to efficiently address shape uncertainty arising from the absence of depth information and intra-class shape variations.


As shown in \cref{tab:camera}, on NOCS-CAMERA25 dataset, LaPose also achieves state-of-the-art performance.
Specifically, we surpass DMSR by 3.0\% and 3.8\% on the strictest metrics $NIoU_{75}$ and $10^\circ 0.2d$ respectively.
As for metric $NIoU_{50}$ and $10^\circ 0.5d$, the performance gap between LaPose and DMSR is 3.4\% and 5.0\%.
In comparison to OLD-Net, LaPose achieves an accuracy of 80.0\% in $10^\circ$ and 45.4\% in $0.2d$, which is 26.7\% and 28.2\% higher than OLD-Net respectively.
These results underscore the efficacy of LaPose across diverse scenarios.

\subsection{Ablation Studies}
\subsubsection{Effect of scale-agnostic pose representation:}
In Tab.~\ref{tab:ablation} (B), we estimate the metric scale and feed the scaled object coordinates to the PnP module to predict the absolute-scale pose. 
In comparison to Tab.~\ref{tab:ablation} (A), where scale-agnostic pose representation is employed, the performance under all metrics drops significantly, underscoring the efficacy of our proposed scale-agnostic pose representation.
Similar trends are also evident in Fig.~\ref{fig:scale-ambiguity} (B).


\subsubsection{Choice of the feature:}
Comparing Tab.~\ref{tab:ablation} (C) to (A) reveals that using only DINOv2 features leads to inferior results.
Additionally, by comparing Tab.~\ref{tab:ablation} (D) to (A) and (C), it is evident that the utilization of both feature streams for predicting NOCS maps enhances performance across all metrics compared to solely relying on a single stream.
Specifically, utilizing two feature streams (Tab.~\ref{tab:ablation} (D)) leads to improvements of 4.6\% in $NIoU_{25}$ and 6.7\% in $10^\circ 0.5d$ respectively, when compared to only using $F_{conv}$ (Tab.~\ref{tab:ablation} (A)).

\subsubsection{Modelling object shape as a probabilistic distribution:}
By comparing \cref{tab:ablation} (E) to (A), we can see that modeling NOCS coordinates as the Laplacian distribution brings 5.2\% and 5.1\% improvements on $NIoU_{25}$ and $10^\circ 0.5d$ metrics respectively.
These improvements are attributed to the predicted variance maps, which guide the PnP module to identify erroneous NOCS predictions.
However, modeling object shape as a Gaussian distribution by using Gaussian aleatoric uncertainty loss as proposed in \cite{gaussianuncertainties}, does not yield positive influences (see \cref{tab:ablation} (F)).
Comparing Tab.~\ref{tab:ablation} (G) and (D), we observe improvements in $NIoU_{50}$ by 5.8\% and $10^\circ 0.2d$ by 4.9\%.
The main reason is that the LMM modeling can capture diverse aspects of object geometry.

\input{tables/ablation}


%% file: tables/norm_result.tex
\begin{table}[t]
    \centering
    \caption{Comparison with state-of-the-art methods on NOCS-REAL275 using scale-agnostic evaluation metrics.}
    \label{tab:norm}
        \tablestyle{5pt}{1.2}
    \begin{tabular}{@{}c|c c c|c c c c c@{}}
    \toprule
        Method & $NIoU_{25}$ & $NIoU_{50}$ & $NIoU_{75}$ & $10^\circ 0.2d$ & $10^\circ 0.5d$ & $0.2d$ & $0.5d$ & $10^\circ$ \\ \hline
        MSOS \cite{lee2021msos} & 36.9 & 9.7 & 0.7 & 3.3 & 15.3 & 10.6 & 50.8 & 17.0 \\ 
        OLD-Net \cite{fan2022oldnet} & 31.5 & 6.2 & 0.1 & 2.8 & 12.2 & 9.0 & 44.0 & 14.8 \\
        DMSR \cite{wei2023dmsr} & 57.2 & 38.4 & 9.7 & 26.0 & 44.9 & 35.8 & 67.2 & 36.9 \\ 
         \hline
        {Ours}  & \textbf{70.7} & 47.9 & 15.8 & 37.4 & \textbf{57.4} & 46.9 & \textbf{78.8} & \textbf{60.7} \\ 
        {Ours (M)} & 66.4 & \textbf{48.8} & \textbf{20.5} & \textbf{39.7} & 55.4 & \textbf{48.6} & 74.9 & 60.2 \\ 
        \bottomrule
    \end{tabular}

\end{table}

%% file: tables/abs_result.tex
\begin{table}[t]
    \centering
    \caption{Comparison with state-of-the-art methods on NOCS-REAL275 using evaluation metrics with absolute object scale.}
    \label{tab:abs}
    \tablestyle{5pt}{1.2}
    \begin{tabular}{@{}c|c c c|c c c c c@{}}
    \toprule
        Method & $IoU_{25}$ & $IoU_{50}$ & $IoU_{75}$ & $10^\circ 10cm$ & $10cm$ \\ \hline
        MSOS \cite{lee2021msos} & 33.2 & 13.6 & 1.0 & 11.8 & 43.4   \\ 
        OLD-Net \cite{fan2022oldnet} & 26.4 & 7.7 & 0.4 & 8.6 & 31.4  \\
        DMSR \cite{wei2023dmsr} & 37.4 & 16.3 & 3.2 & 25.2 & 40.0   \\ 
         \hline
        Ours  & \textbf{41.2} & 17.5 & 2.6 & \textbf{30.5} & \textbf{44.4}  \\ 
        Ours (M) & 40.2 & \textbf{18.3} & \textbf{4.1} & 27.7 & 43.7 \\ 
        \bottomrule
    \end{tabular}
\end{table}

%% file: figs/cat-result-ddpnp.tex
\begin{figure}[t]
\centering
\includegraphics[width=0.8\textwidth]{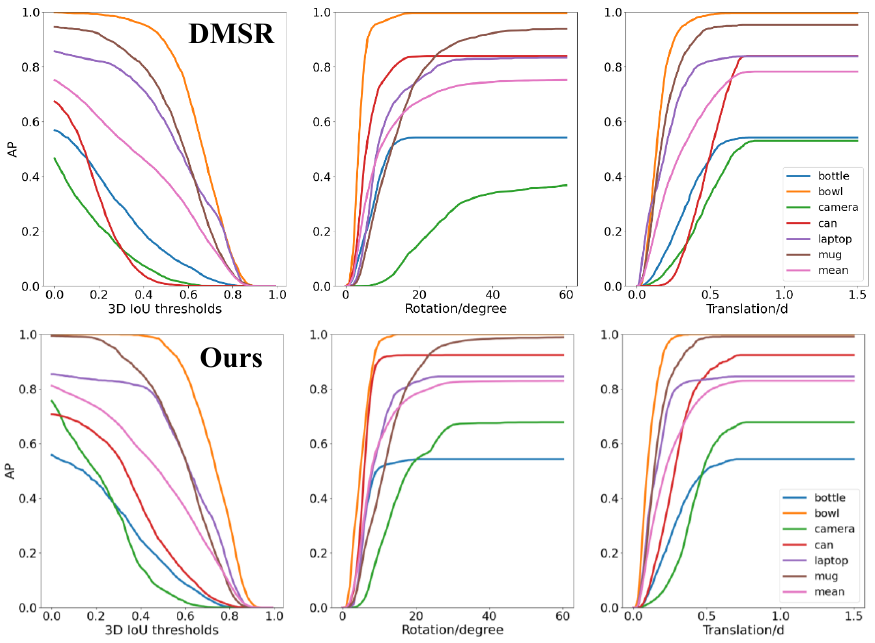}
\caption{The Average Precision (AP) under different thresholds on scale-agnostic 3D IoU, rotation, and translation of DMSR and \textbf{Ours}. Best viewed in color and zoom-in.
}
\label{fig:category_curve}
\end{figure}

%% file: tables/camera_result.tex
\begin{table}[t]
    \centering
    \caption{Comparison with state-of-the-art methods on NOCS-CAMERA25  using scale-agnostic evaluation metrics.}
    \label{tab:camera}
    \tablestyle{4pt}{1.2}
    \begin{tabular}{@{}c|c c c|c c c c c@{}}
    \toprule
        Method & $NIoU_{25}$ & $NIoU_{50}$ & $NIoU_{75}$ & $10^\circ 0.2d$ & $10^\circ 0.5d$ & $10^\circ$ & $0.2d$ & $0.5d$ \\ \hline
        MSOS \cite{lee2021msos} & 35.1 & 9.9 & 0.8 & 5.9 & 31.6 & 48.6 & 8.9 & 47.2 \\ 
        OLD-Net \cite{fan2022oldnet} & 50.4 & 14.5 & 0.5 & 11.3 & 39.9 & 53.3 & 17.2 & 60.5 \\
        DMSR \cite{wei2023dmsr} & 74.4 & 46.0 & 11.1 & 38.6 & 68.1 & 74.4 & 42.8 & 79.9 \\ 
         \hline
        Ours & \textbf{75.3} & \textbf{49.4} & \textbf{14.1} & \textbf{42.4} & \textbf{73.1} & \textbf{80.0} & \textbf{45.4} & \textbf{81.2} \\ 
        \bottomrule
    \end{tabular}
\end{table}

%% file: tables/ablation.tex
\begin{table}[!t]
    \centering
    \caption{Ablation studies on NOCS-REAL275. $\mathcal{F}_{conv}$ and $\mathcal{F}_{dino}$ denote the semantic feature extracted by the convolutional network and DINOv2. 
    $\sigma^2$ denotes whether variances are predicted and fed into the PnP module.
    If $\sigma^2$ is used, the modeling distribution is either Gaussian (Gaus.) or Laplacian (Lap.).
    SAP denotes whether the scale-agnostic pose representation is used.
    }
    \label{tab:ablation}
    \tablestyle{5pt}{1.2}
    \begin{tabular}{@{}c|ccc|c|cc|cc@{}}
    \toprule
        No. & $\mathcal{F}_{conv}$ & $\mathcal{F}_{dino}$ & $\sigma^2$ & SAP  & $NIoU_{25}$ & $NIoU_{50}$ & $10^\circ 0.2d$ & $10^\circ 0.5d$ \\ \hline
        A & \checkmark & ~  & ~ & \checkmark  &  60.3 & 39.7 & 29.0 & 46.1 \\ 
        B & \checkmark & ~  & ~ & ~  & 37.8 & 17.6 & 10.5 & 33.2 \\ 
        \hline
        C &  & \checkmark & ~  & \checkmark  & 61.7 & 26.9 & 16.9 & 38.5 \\ 
        D & \checkmark & \checkmark  &  & \checkmark & 64.9 & 42.1 & 32.5 & 52.8 \\
        \hline
        E & \checkmark &  & Lap.  & \checkmark  & 65.5 & 43.4 & 33.6 & 51.2 \\ 
        F & \checkmark &  & Gaus.  & \checkmark  & 59.1 & 37.6 & 28.3 & 44.7 \\ 
        \hline
        G & \checkmark & \checkmark  & Lap. & \checkmark & \textbf{70.7} & \textbf{47.9} & \textbf{37.4} & \textbf{57.4} \\
\bottomrule
    \end{tabular}
\end{table}

%% file: Sections/5-conclusion.tex
\section{Conclusion}


In this paper, we propose LaPose, a novel framework that models the object shape as the Laplacian Mixture Model (LMM) for RGB-based category-level object pose estimation.
Specifically, we integrate two independent Laplacian distributions derived from two different feature streams.
We establish 2D-3D correspondences using the estimated LMM and solve the pose via a PnP module. 
Our proposed scale-agnostic pose representation effectively addresses scale ambiguity and ensures stable and efficient training. 
Extensive experiments on NOCS datasets demonstrate that LaPose achieves state-of-the-art performance. 
Looking ahead, we plan to extend LaPose to applications in robotic manipulation.
Limitations are discussed in Sup. Mat.

\section*{Acknowledgement}
This work was supported by the National Key R\&D Program of China under Grant 2018AAA0102801.

%% file: _main.bbl
\begin{thebibliography}{10}
\providecommand{\url}[1]{\texttt{#1}}
\providecommand{\urlprefix}{URL }
\providecommand{\doi}[1]{https://doi.org/#1}

\bibitem{amir2022dino}
Amir, S., Gandelsman, Y., Bagon, S., Dekel, T.: Deep vit features as dense visual descriptors. arXiv preprint arXiv:2112.05814  \textbf{2}(3), ~4 (2021)

\bibitem{cass}
Chen, D., Li, J., Wang, Z., Xu, K.: Learning canonical shape space for category-level 6d object pose and size estimation. In: CVPR. pp. 11973--11982 (2020)

\bibitem{chen2022epro}
Chen, H., Wang, P., Wang, F., Tian, W., Xiong, L., Li, H.: Epro-pnp: Generalized end-to-end probabilistic perspective-n-points for monocular object pose estimation. In: Proceedings of the IEEE/CVF Conference on Computer Vision and Pattern Recognition. pp. 2781--2790 (2022)

\bibitem{sgpa}
Chen, K., Dou, Q.: Sgpa: Structure-guided prior adaptation for category-level 6d object pose estimation. In: ICCV. pp. 2773--2782 (2021)

\bibitem{fs-net}
Chen, W., Jia, X., Chang, H.J., Duan, J., Linlin, S., Leonardis, A.: Fs-net: Fast shape-based network for category-level 6d object pose estimation with decoupled rotation mechanism. In: CVPR. pp. 1581--1590 (June 2021)

\bibitem{chen2020rgbsyn}
Chen, X., Dong, Z., Song, J., Geiger, A., Hilliges, O.: Category level object pose estimation via neural analysis-by-synthesis. In: ECCV. pp. 139--156. Springer (2020)

\bibitem{chen2023secondpose}
Chen, Y., Di, Y., Zhai, G., Manhardt, F., Zhang, C., Zhang, R., Tombari, F., Navab, N., Busam, B.: Secondpose: Se (3)-consistent dual-stream feature fusion for category-level pose estimation. arXiv preprint arXiv:2311.11125  (2023)

\bibitem{chen2020monopair}
Chen, Y., Tai, L., Sun, K., Li, M.: Monopair: Monocular 3d object detection using pairwise spatial relationships. In: CVPR. pp. 12093--12102 (2020)

\bibitem{robotics}
Deng, X., Xiang, Y., Mousavian, A., Eppner, C., Bretl, T., Fox, D.: Self-supervised 6d object pose estimation for robot manipulation. In: ICRA. pp. 3665--3671. IEEE (2020)

\bibitem{sopose}
Di, Y., Manhardt, F., Wang, G., Ji, X., Navab, N., Tombari, F.: So-pose: Exploiting self-occlusion for direct 6d pose estimation. In: ICCV. pp. 12396--12405 (2021)

\bibitem{GPV-Pose}
Di, Y., Zhang, R., Lou, Z., Manhardt, F., Ji, X., Navab, N., Tombari, F.: Gpv-pose: Category-level object pose estimation via geometry-guided point-wise voting. arXiv preprint  (2022)

\bibitem{fan2022oldnet}
Fan, Z., Song, Z., Xu, J., Wang, Z., Wu, K., Liu, H., He, J.: Object level depth reconstruction for category level 6d object pose estimation from monocular rgb image. In: ECCV. pp. 220--236. Springer (2022)

\bibitem{ACR-Pose}
Fan, Z., Song, Z., Xu, J., Wang, Z., Wu, K., Liu, H., He, J.: Acr-pose: Adversarial canonical representation reconstruction network for category level 6d object pose estimation. arXiv preprint arXiv:2111.10524  (2021)

\bibitem{maskrcnn}
He, K., Gkioxari, G., Doll{\'a}r, P., Girshick, R.: Mask r-cnn. In: ICCV. pp. 2961--2969 (2017)

\bibitem{FFB6D}
He, Y., Huang, H., Fan, H., Chen, Q., Sun, J.: Ffb6d: A full flow bidirectional fusion network for 6d pose estimation. In: CVPR. pp. 3003--3013 (June 2021)

\bibitem{pvn3d}
He, Y., Sun, W., Huang, H., Liu, J., Fan, H., Sun, J.: Pvn3d: A deep point-wise 3d keypoints voting network for 6dof pose estimation. In: CVPR. pp. 11632--11641 (2020)

\bibitem{hodan2020epos}
Hodan, T., Barath, D., Matas, J.: Epos: Estimating 6d pose of objects with symmetries. In: CVPR. pp. 11703--11712 (2020)

\bibitem{mobilenetv3}
Howard, A., Sandler, M., Chen, B., Wang, W., Chen, L.C., Tan, M., Chu, G., Vasudevan, V., Zhu, Y., Pang, R., Adam, H., Le, Q.: Searching for mobilenetv3. In: 2019 IEEE/CVF International Conference on Computer Vision (ICCV). pp. 1314--1324 (2019)

\bibitem{hu2022perspective}
Hu, Y., Fua, P., Salzmann, M.: Perspective flow aggregation for data-limited 6d object pose estimation. In: European Conference on Computer Vision. pp. 89--106. Springer (2022)

\bibitem{kehl2017ssd}
Kehl, W., Manhardt, F., Tombari, F., Ilic, S., Navab, N.: Ssd-6d: Making rgb-based 3d detection and 6d pose estimation great again. In: ICCV. pp. 1521--1529 (2017)

\bibitem{kehl2016deep}
Kehl, W., Milletari, F., Tombari, F., Ilic, S., Navab, N.: Deep learning of local rgb-d patches for 3d object detection and 6d pose estimation. In: ECCV (2016)

\bibitem{gaussianuncertainties}
Kendall, A., Gal, Y.: What uncertainties do we need in bayesian deep learning for computer vision? NeurIPS  \textbf{30} (2017)

\bibitem{labbe2020cosypose}
Labb{\'e}, Y., Carpentier, J., Aubry, M., Sivic, J.: Cosypose: Consistent multi-view multi-object 6d pose estimation. In: ECCV. pp. 574--591. Springer (2020)

\bibitem{lee2021msos}
Lee, T., Lee, B.U., Kim, M., Kweon, I.S.: Category-level metric scale object shape and pose estimation. IEEE RA-L  \textbf{6}(4),  8575--8582 (2021)

\bibitem{lepetit2009epnp}
Lepetit, V., Moreno-Noguer, F., Fua, P.: Epnp: An accurate o (n) solution to the pnp problem. International journal of computer vision  \textbf{81}(2), ~155 (2009)

\bibitem{li2019deepim}
Li, Y., Wang, G., Ji, X., Xiang, Y., Fox, D.: {DeepIM}: Deep iterative matching for 6d pose estimation. IJCV pp. 1--22 (2019)

\bibitem{li2019cdpn}
Li, Z., Wang, G., Ji, X.: {CDPN}: {C}oordinates-{B}ased {D}isentangled {P}ose {N}etwork for {R}eal-{T}ime {RGB}-{B}ased 6-{DoF} {O}bject {P}ose {E}stimation. In: ICCV. pp. 7678--7687 (2019)

\bibitem{donet}
Lin, H., Liu, Z., Cheang, C., Zhang, L., Fu, Y., Xue, X.: Donet: Learning category-level 6d object pose and size estimation from depth observation. arXiv preprint arXiv:2106.14193  (2021)

\bibitem{lin2022dpdn}
Lin, J., Wei, Z., Ding, C., Jia, K.: Category-level 6d object pose and size estimation using self-supervised deep prior deformation networks. In: ECCV. pp. 19--34. Springer (2022)

\bibitem{dualposenet}
Lin, J., Wei, Z., Li, Z., Xu, S., Jia, K., Li, Y.: Dualposenet: Category-level 6d object pose and size estimation using dual pose network with refined learning of pose consistency. arXiv preprint arXiv:2103.06526  (2021)

\bibitem{lin2023vinet}
Lin, J., Wei, Z., Zhang, Y., Jia, K.: Vi-net: Boosting category-level 6d object pose estimation via learning decoupled rotations on the spherical representations. In: ICCV. pp. 14001--14011 (2023)

\bibitem{lipson2022coupled}
Lipson, L., Teed, Z., Goyal, A., Deng, J.: Coupled iterative refinement for 6d multi-object pose estimation. In: CVPR. pp. 6728--6737 (2022)

\bibitem{liu2023istnet}
Liu, J., Chen, Y., Ye, X., Qi, X.: Ist-net: Prior-free category-level pose estimation with implicit space transformation. In: ICCV. pp. 13978--13988 (2023)

\bibitem{ranger1}
Liu, L., Jiang, H., He, P., Chen, W., Liu, X., Gao, J., Han, J.: On the variance of the adaptive learning rate and beyond. In: ICLR (2019)

\bibitem{liu2022convnext}
Liu, Z., Mao, H., Wu, C.Y., Feichtenhofer, C., Darrell, T., Xie, S.: A convnet for the 2020s. In: CVPR. pp. 11976--11986 (2022)

\bibitem{nie2020total3dunderstanding}
Nie, Y., Han, X., Guo, S., Zheng, Y., Chang, J., Zhang, J.J.: Total3dunderstanding: Joint layout, object pose and mesh reconstruction for indoor scenes from a single image. In: CVPR. pp. 55--64 (2020)

\bibitem{oquab2023dinov2}
Oquab, M., Darcet, T., Moutakanni, T., Vo, H.V., Szafraniec, M., Khalidov, V., Fernandez, P., HAZIZA, D., Massa, F., El-Nouby, A., Assran, M., Ballas, N., Galuba, W., Howes, R., Huang, P.Y., Li, S.W., Misra, I., Rabbat, M., Sharma, V., Synnaeve, G., Xu, H., Jegou, H., Mairal, J., Labatut, P., Joulin, A., Bojanowski, P.: {DINO}v2: Learning robust visual features without supervision. Transactions on Machine Learning Research  (2024)

\bibitem{park2019pix2pose}
Park, K., Patten, T., Vincze, M.: Pix2pose: Pixel-wise coordinate regression of objects for 6d pose estimation. In: ICCV (2019)

\bibitem{peng2019pvnet}
Peng, S., Liu, Y., Huang, Q., Zhou, X., Bao, H.: Pvnet: Pixel-wise voting network for 6dof pose estimation. In: CVPR (2019)

\bibitem{Ranftl_2021_ICCV_dpt}
Ranftl, R., Bochkovskiy, A., Koltun, V.: Vision transformers for dense prediction. In: ICCV. pp. 12179--12188 (October 2021)

\bibitem{hybridpose}
Song, C., Song, J., Huang, Q.: Hybridpose: 6d object pose estimation under hybrid representations. In: CVPR. pp. 431--440 (2020)

\bibitem{arvr}
Su, Y., Rambach, J., Minaskan, N., Lesur, P., Pagani, A., Stricker, D.: Deep multi-state object pose estimation for augmented reality assembly. In: 2019 IEEE International Symposium on Mixed and Augmented Reality Adjunct (ISMAR-Adjunct). pp. 222--227. IEEE (2019)

\bibitem{su2022zebrapose}
Su, Y., Saleh, M., Fetzer, T., Rambach, J., Navab, N., Busam, B., Stricker, D., Tombari, F.: Zebrapose: Coarse to fine surface encoding for 6dof object pose estimation. In: CVPR. pp. 6738--6748 (2022)

\bibitem{shape_deform}
Tian, M., Ang, M.H., Lee, G.H.: Shape prior deformation for categorical 6d object pose and size estimation. In: ECCV. pp. 530--546. Springer (2020)

\bibitem{umeyama}
Umeyama, S.: Least-squares estimation of transformation parameters between two point patterns. IEEE TPAMI  \textbf{13}(04),  376--380 (1991). \doi{10.1109/34.88573}

\bibitem{wang2019densefusion}
Wang, C., Xu, D., Zhu, Y., Mart{\'\i}n-Mart{\'\i}n, R., Lu, C., Fei-Fei, L., Savarese, S.: {DenseFusion}: 6d object pose estimation by iterative dense fusion. In: CVPR. pp. 3343--3352 (2019)

\bibitem{GDRN}
Wang, G., Manhardt, F., Tombari, F., Ji, X.: Gdr-net: Geometry-guided direct regression network for monocular 6d object pose estimation. In: CVPR (June 2021)

\bibitem{NOCS}
Wang, H., Sridhar, S., Huang, J., Valentin, J., Song, S., Guibas, L.J.: Normalized object coordinate space for category-level 6d object pose and size estimation. In: CVPR. pp. 2642--2651 (2019)

\bibitem{wei2023dmsr}
Wei, J., Song, X., Liu, W., Kneip, L., Li, H., Ji, P.: Rgb-based category-level object pose estimation via decoupled metric scale recovery. arXiv preprint arXiv:2309.10255  (2023)

\bibitem{Wohlhart2015b}
Wohlhart, P., Lepetit, V.: {Learning descriptors for object recognition and 3D pose estimation}. CVPR pp. 3109--3118 (2015). \doi{10.1109/CVPR.2015.7298930}

\bibitem{xiang2017posecnn}
Xiang, Y., Schmidt, T., Narayanan, V., Fox, D.: {PoseCNN}: A convolutional neural network for 6{D} object pose estimation in cluttered scenes. RSS  (2018)

\bibitem{ranger3}
Yong, H., Huang, J., Hua, X., Zhang, L.: Gradient centralization: A new optimization technique for deep neural networks. In: ECCV. pp. 635--652. Springer (2020)

\bibitem{zakharov2019dpod}
Zakharov, S., Shugurov, I., Ilic, S.: Dpod: Dense 6d pose object detector in rgb images. In: ICCV (2019)

\bibitem{zhang2023tale}
Zhang, J., Herrmann, C., Hur, J., Cabrera, L.P., Jampani, V., Sun, D., Yang, M.H.: A tale of two features: Stable diffusion complements dino for zero-shot semantic correspondence. arXiv preprint arXiv:2305.15347  (2023)

\bibitem{ranger2}
Zhang, M., Lucas, J., Ba, J., Hinton, G.E.: Lookahead optimizer: k steps forward, 1 step back. In: Wallach, H., Larochelle, H., Beygelzimer, A., d\textquotesingle Alch\'{e}-Buc, F., Fox, E., Garnett, R. (eds.) NeurIPS. vol.~32. Curran Associates, Inc. (2019)

\bibitem{zhang2022rbp}
Zhang, R., Di, Y., Lou, Z., Manhardt, F., Tombari, F., Ji, X.: Rbp-pose: Residual bounding box projection for category-level pose estimation. In: ECCV. pp. 655--672. Springer (2022)

\bibitem{zhang2022ssp}
Zhang, R., Di, Y., Manhardt, F., Tombari, F., Ji, X.: Ssp-pose: Symmetry-aware shape prior deformation for direct category-level object pose estimation. In: IROS. pp. 7452--7459. IEEE (2022)

\bibitem{zheng2023hs}
Zheng, L., Wang, C., Sun, Y., Dasgupta, E., Chen, H., Leonardis, A., Zhang, W., Chang, H.J.: Hs-pose: Hybrid scope feature extraction for category-level object pose estimation. In: CVPR. pp. 17163--17173 (2023)

\end{thebibliography}
